\documentclass[english]{article}

\usepackage{hyperref}
\usepackage{url}
\usepackage{graphicx} 
\usepackage{subfigure}
\usepackage{multirow}
\usepackage{amsthm}
\usepackage{amsmath,amssymb,bm}
\usepackage[ruled] {algorithm2e}
\usepackage{graphicx}
\usepackage{wrapfig}

\newtheorem{theorem}{Theorem}

\usepackage{wrapfig}
\usepackage{lipsum}

\author{
{\sc Chu Wang}
\thanks{Program in Applied and Computational Mathematics,
       Princeton University, 
{\tt   chuw}@{\tt math.princeton.edu }}
\and
  {\sc Yingfei Wang}
\thanks{Department of Computer Science,
       Princeton University, 
{\tt yingfei}@{\tt cs.princeton.edu }}
\and
{\sc Weinan E}
\thanks{Department of Mathematics,
       Princeton University, 
{\tt   weinan}@{\tt math.princeton.edu }}
\and
{\sc Robert Schapire}
\thanks{Microsoft Research, NY, 
{\tt   schapire}@{\tt microsoft.com }}
}

\title{Functional Frank-Wolfe Boosting for General Loss Functions}

\begin{document}
\date{}

\maketitle

\begin{abstract}
Boosting is a generic learning method for classification and regression. Yet, as the number of base hypotheses becomes larger,  boosting can lead to a deterioration of test performance.  Overfitting is an important and ubiquitous phenomenon, especially in regression settings. To avoid overfitting, we consider using $l_1$ regularization. We propose a novel Frank-Wolfe type boosting algorithm (FWBoost) applied to general  loss functions.  By using exponential loss, the FWBoost algorithm can be rewritten as a variant of  AdaBoost  for binary classification.  FWBoost algorithms have exactly the same form as existing boosting methods, in terms of making calls to a base learning algorithm with different weights update.  This  direct connection between boosting and Frank-Wolfe yields a new algorithm that is as practical as existing boosting methods but with new guarantees and rates of convergence. Experimental results   show that the test performance of FWBoost is not degraded with larger rounds in boosting, which is consistent with the theoretical analysis.
\end{abstract}

\section{Introduction}
Boosting is a generic learning method for classification and regression.  From a statistical perspective, boosting is often viewed as a method for empirical minimization of an appropriate loss function in a greedy fashion \cite{breiman1998arcing,friedman2001greedy,mason1999functional,friedman2000additive}. It is designed to iteratively select a base hypothesis $h$ that leads to the largest reduction of empirical risk at each step from the base hypothesis space $\mathcal{H}$.  The family of combined hypotheses $F(\cdot)$ being considered is the set of ensembles of base hypotheses $F=\sum_{j}\alpha_jh_j$ with $h_j \in \mathcal{H}$ and $\alpha_j \in \mathbb{R}.$
Boosting procedures have drawn much attention in the machine leaning and statistics community due to their superior empirical performance ever since the first practical boosting algorithm, Adaboost \cite{freund1997decision}. Many other generalizations and extensions have since been proposed \cite{friedman2001greedy,friedman2000additive,schapire2003boosting,breiman1998arcing}.

For classification, the margin theory \cite{schapire1998boosting}  explains AdaBoost's resistance to overfitting, although it can overfit when the base hypotheses are too complex relative to the  size of the training set  or when the learning algorithm is  unable to achieve large margins. Yet all this theory is not applicable for regression and overfitting is ubiquitous in regression settings \cite{grove1998boosting}.  Various methods have been proposed to  avoid overfitting. For example, the early stopping framework  has been  considered for different  loss functions  \cite{buhlmann2002consistency,zhang2005boosting}.   Friedman   \cite{friedman2002stochastic} proposed stochastic gradient boosting.  Duchi and Singer  \cite{duchi2009boosting}  studied penalties for Adaboost based on the $l_1$, $l_2$, and $l_\infty$ norms of the predictors.  A different method to avoid overfitting (and obtain rates of convergence) is through restricting the $l_1$ regularization of the weights of the composite base hypotheses. For classification problems,  this point of view is taken up in \cite{lugosi2004bayes,xi2009speed,shen2010totally}.  As an alternative,  shrinkage has been applied \cite{hastie2009elements}.  It is surprising to find that there are fewer $l_1$ regularized boosting algorithms for regression problems other than shrinkage. There are extensive studies on minimizing the Lasso loss function ($l_1$ penalized loss) \cite{tibshirani1996regression,osborne2000lasso,shalev2010trading}. Yet most of the studies cannot be applied to general loss functions as in ensemble learning (a.k.a, boosting) with predictors in the form of $F=\sum_{j}\alpha_jh_j$. Zhao and Yu  \cite{zhao2004boosted} proposed the Boosted Lasso (BLasso) algorithm which ties the boosting algorithm with the $l_1$ penalized  Lasso method with an emphasis on tracing the regularization paths.

In this paper, we study algorithms for loss minimization subject to an explicit constraint on the $l_1$ norm of the weights. This approach yields a bound on the generalization error that depends only on the constraint and that is independent of the number of iterations. In Section  \ref{sec:GLF}, we propose a novel Frank-Wolfe type boosting algorithm (FWBoost) for  general loss functions. By using exponential loss for binary classification, the FWBoost algorithm can be reduced to an AdaBoost-like algorithm  (AdaBoost.FW) derived as a coordinate descent method in Section \ref{sec:Class}. By making a direct connection between boosting and Frank-Wolfe, the FWBoost algorithms   have exactly the same form as existing boosting methods with the same number of calls to a base learners but with new guarantees and rate of convergence $O(\frac{1}{t})$.  Experimental results  in Section \ref{sec:exp} show that the test performance of FWBoost is not degraded with larger rounds in the boosting, which is consistent with the theoretical analysis in Section \ref{sec:GLF}.

\section{Preliminaries}\label{sec:pre}
In this section, we briefly overview the gradient boosting algorithm and Frank-Wolfe algorithms.
\paragraph{Gradient Boosting}
We assume that the samples $(\bm{x},y)$ are independently chosen from the same but unknown distribution $\mathcal{D}$ where $\bm{x}$ is an instance from domain $\mathcal{X}$  and the univariate $y \in \mathcal{Y}$ can be continuous (regression problem) or discrete (classification problem).
During training, a learning algorithm receives a training set $S = \{(\bm{x}_1, y_1), (\bm{x}_2, y_2), \cdots, (\bm{x}_m, y_m)\}$. 
The goal is to estimate the function/hypothsis $F:\mathcal{X} \rightarrow \mathcal{Y}$ that minimizes the expected value of some specified loss function $l(F(\bm{x}),y)$:
$
\arg\min_{F}\mathbb{E}_{\bm{x},y}[l(F(\bm{x}), y)].
$
The hypothesis space being considered is the set of ensembles $\mathcal{F}$ of base hypothesis $h_j$ from the base hypothesis space $\mathcal{H}$:
$$\mathcal{F}=\Big{\{}F(\bm{x}):F(\bm{x})=\sum_j\alpha_jh_j(\bm{x}), \bm{x}\in \mathcal{X}, \alpha_j\in\mathbb{R}, h_j\in\mathcal{H}\Big{\}}.$$

Gradient boosting \cite{friedman2001greedy,mason1999functional} tries to find an approximation $\hat{F}(\bm{x})$ that minimizes the empirical risk $L(F,\bm{y})=\frac{1}{m}\sum_{i=1}^ml(F(\bm{x}_i),y_i)$ on the training set. 
It does so by iteratively building up the solution in a greedy fashion:
$$
(\alpha_t, h_t)=\arg\min_{\alpha, h }L(F_{t-1}(\bm{x}_i)+\alpha h(\bm{x}_i),y_i), ~~
F_t(\bm{x})=F_{t-1}(\bm{x})+\alpha_th_t(\bm{x}).
$$

 \begin{wrapfigure}{L}{0.41\textwidth}
  	\begin{minipage}{0.41\textwidth}
	\begin{algorithm}[H]\label{FW}
	\caption{Frank-Wolfe}
	\SetKwData{Left}{left}\SetKwData{This}{this}\SetKwData{Up}{up}
	\SetKwFunction{Union}{Union}\SetKwFunction{FindCompress}{FindCompress}
	\SetKwInOut{Input}{input}\SetKwInOut{Output}{output}
	 \SetAlgoLined
	Let $\bm{x}^{(0)} \in \mathcal{D}$ \\
 	\For{$t=1$ to $T$}{
	$\bm{s} := \arg\min_{\bm{s} \in \mathcal{D}} \langle \bm{s}, \nabla f(\bm{x}^{(t)})\rangle$\\
	$\gamma_t := \frac{2}{t+2}$\\
	$\bm{x}^{(t+1)} := (1-\gamma_t)\bm{x}^{(t)}+\gamma_t \bm{s}$\\
	}
	\end{algorithm}
	\end{minipage} 
\end{wrapfigure}
However, there is no simply way to exactly solve the problem of choosing at each step the best $h$ for an arbitrary loss function $l$ since the base hypothesis space is usually large or infinite.  A typical strategy is by applying functional gradient descent. For the purpose of minimizing the empirical risk, we only care about the value of $F$ at $\bm{x}_1,...,\bm{x}_m$. 
Thus we can view $L(F,\bm{y})$ as a function of a vector of values $( F(\bm{x}_1),F(\bm{x}_2),...,F(\bm{x}_m) )$ and then  calculate the negative gradient of $L$: $r_{t,i}=-\big[\frac{\partial L(F, \bm{y})}{\partial F(\bm{x}_i)}\big]_{F=F_{t-1}} $.  A base hypothesis $h$ is chosen to most closely approximates the negative gradient and a step size is found by line search.

\paragraph{Frank-Wolfe Algorithms.}

Frank-Wolfe algorithm \cite{frank1956algorithm} (also known as conditional gradient method) 
is one of the simplest and popular iterative first-order method for general constrained optimization problems. 
Given an continuously differentiable function $f$ and a compact convex domain $\mathcal{D}$,
the objective is 
\begin{equation}\label{obj}
\min_{\bm{x} \in \mathcal{D}} f(\bm{x}).
\end{equation}
At each iteration, the algorithm searches for the minimizer of the linearized version of the optimization problem under the same constraint and then applies the descend method along the direction from the current position towards this minimizer. By confining the step size $\gamma$ in $[0,1]$, this algorithm  automatically maintains the current position $\bm{x}$ in the feasible region. It is known that the convergence of Algorithm \ref{FW} satisfies $f(\bm{x}^{(t)})-f(\bm{x}^*) \le O(\frac{1}{t})$, where $x^*$ is the solution to \eqref{obj} \cite{frank1956algorithm,jaggi2013revisiting}.

There are several variants of the classic Frank-Wolfe algorithm, including line-search variant and the fully corrective variant, see \cite{jaggi2013revisiting} for a recent overview.
Another important variant is the use of away-steps, as  described in  \cite{guelat1986some}.  
The idea is that in each iteration, we may potentially remove  an old atom with  bad performance instead of adding a new one, which can improve the sparsity of the iterates \cite{clarkson2010coresets}.

\section{Functional Frank-Wolfe Boosting for General Loss Functions}\label{sec:GLF}
In order to prevent boosting algorithms from overfitting, we consider the  constraint version of the $l_1$ regularized empirical risk minimization:
\begin{equation}\label{alpha}
\min_{\|\bm{\alpha}\|_1\le C}L(F,\bm{y}).
\end{equation}
We will use $\mathcal{H}^D$ to denote the subset of $\mathcal{H}$ whose $l_\infty$ norm is bounded by $D$. 
Bounding the $l_\infty$ norm of the base hypotheses is essential because otherwise if $\mathcal{H}$ is closed under scalar multiplication, we could always shrink $\alpha$ and enlarge $h$ to remove the effect of the constraint $\|\boldsymbol{\alpha}\|_1\le C$.
It is clear that $D$ and $C$ are two dual parameters and the intrinsic degree of freedom is 1. 
Thus without loss of generality we  consider the base hypotheses in $\mathcal{H}^1$.

Since the base hypothesis space is usually large or infinite, we take a functional view by treating the empirical risk function $L$ as a functional which takes as input  another function $F$ rather than viewing our objective function, as in Eq. \eqref{alpha}, as a function of a set parameters $\alpha_1, \alpha_2,\dots, $ representing the weights over all the base hypothesis. Define the hypotheses space $\mathcal{F}^C$ as
\begin{equation*}
\mathcal{F}^C=\big{\{}F:F(\bm{x})=\sum_{j}\alpha_jh_j(\bm{x}), h_j \in \mathcal{H}^1, \|\bm{\alpha}\|_1\le C, \alpha_j \in \mathbb{R}\big{\}}.
\end{equation*} We use $\mathcal{F}_T^C$ to denote the hypothesis space after $T$-step boosting.

The $l_1$ regularized empirical risk minimization can be written analogously to Eq. \eqref{obj} in functional space  with compact  convex domain $\mathcal{F}^C$ as:
\begin{equation}\label{Fobj}
\min_{F \in \mathcal{F}^C}L(F,\bm{y}).
\end{equation}
 
We will see that this perspective overcomes possible computational difficulties encountered by optimizing over the set of parameters $\alpha_1, ..., \alpha_m$ in coordinate descent for $l_1$ regularized empirical risk minimization and is widely applicable to general loss functions. In fact, by applying Frank-Wolfe algorithms in functional space, it allows the $l_1$ regularized minimization of many loss functions to be reduced to a sequence of ordinary classification problems or least-squares regression problems.

\subsection{Functional Frank-Wolfe}
For the purpose of optimizing Eq. \eqref{Fobj}, we only care about the value of $F$ at $\bm{x}_1,...,\bm{x}_m$. Thus we can view $L$ as a function of a vector of values $[ F(\bm{x}_1),F(\bm{x}_2),...,F(\bm{x}_m) ]$.
To simplify notation, we use $f$ and  the vector $[f(\bm{x}_1),f(\bm{x}_2),\dots,f(\bm{x}_m)]$  interchangeably for any function $f$ if no confusion is presented. By analogy of $L$, $F$ and $\mathcal{F}^C$ in Eq. \eqref{Fobj} to $f$, $\bm{x}$ and $\mathcal{D}$ in Eq.\eqref{obj}, respectively, we derive the  basic Frank-Wolfe boosting (FWBoost) framework in Algorithm \ref{FWB}.

\begin{algorithm}[H]
\caption{FWBoost: a generic functional Frank-Wolfe algorithm} \label{FWB}
\SetKwData{Left}{left}\SetKwData{This}{this}\SetKwData{Up}{up}
\SetKwFunction{Union}{Union}\SetKwFunction{FindCompress}{FindCompress}
\SetKwInOut{Input}{input}\SetKwInOut{Output}{output}
 \SetAlgoLined

 \Input{$m$ examples $(\bm{x}_1,y_1),(\bm{x}_2,y_2),\dots(\bm{x}_m,y_m)$ and constant $C>0$}
  \BlankLine

 \For{$t=1$ to $T$}{
~~~~~~Calculate the negative gradient $\bm{r}_t$: $r_{t,i}=-\big[\frac{\partial L(F, \bm{y})}{\partial F(\bm{x}_i)}\big]_{F(\bm{x})=F_{t-1}(\bm{x})} $\\
(a)~~Solve subproblem $h_t=C\cdot\arg\max_{h\in\mathcal{H}^1}\langle h,\bm{r}_t\rangle$, $\gamma_t=2/(t+2)$ \\
~~~~~~$F_t(\bm{x})=F_{t-1}(\bm{x})+\gamma_t\big{(}h_t(\bm{x})-F_{t-1}(\bm{x})\big{)}$
}
\Output{$F_T(x)$}
\end{algorithm}

Let $\alpha_{t,k}$  denote the parameter of $s_k$ at step $t$, $k \le t$. From step $t-1$ to $t$, the updates on $\alpha_{t,k}$ is $\alpha_{t,k}=(1-\gamma_t)\alpha_{t-1,k}$ for $1\le k\le t-1$ and $\alpha_{t,t}=\gamma_tC$. If $\|\bm{\alpha}_{t-1}\|_1\le C$,
 we have $\|\bm{\alpha}_t\|\le (1-\gamma_t)\|\bm{\alpha}_{t-1}\|_1+\gamma_tC\le C$.
Since  $\|\alpha_{1,1}\|_1\le C$,  it is clear that $F_t$ is always in $\mathcal{F}^C$. 


Recall that in the original Frank-Wolfe algorithm, the linearized subproblem $\bm{s}=\arg\min_{\mathbf{s}\in\mathcal{D}}\langle \bm{s}, \nabla f(\bm{x}^{(t)})\rangle$ is in the whole feasible region $\mathcal{D}$ which corresponds to $F_C$ in the learning setting. Here we are able to use $\mathcal{H}^1$  instead of $F_C$in line (a). 
This is because the feasible region is determined by linear constraints and the objective function $\langle h,\bm{r}_t\rangle$ is also linear.
Therefore $h$ attains the maximum at the vertices, which is exactly the function space $C\mathcal{H}^1$.
The way of choosing step length $\gamma_t$ has several variants like line search or total corrective as discussed in Section \ref{sec:pre}.

\subsection{Using Classification and Regression for General Loss Functions}
To solve the central subproblem $\arg\max_{h\in\mathcal{H}^1}\langle h,\bm{r}_t\rangle$, we need to find a function $h$ in $\mathcal{H}^1$ which, as a vector, has the most similar orientation to $\bm{r}$.  If each $h \in \mathcal{H}$ is constrained to take values in $\{-1, +1\}$, the  problem of  maximizing $\langle h,\bm{r}_t\rangle$ on each round is equivalent to an ordinary classification problem. 
To be more specific, let $\tilde{r}_{t,i}=\mathrm{sign}(r_{t,i}) \in \{-1, +1\}$ and $w_t(i)=|r_{t,i}|/\|\bm{r}_t\|_1$.
Then by a standard technique, we have
\begin{eqnarray*}
\arg\max_{h\in\mathcal{H}^1}\langle h,\bm{r}_t\rangle&\propto&
\arg\max_{h\in\mathcal{H}^1}\sum_{i=1}^mh(\bm{x_i})\tilde{r}_{t,i}w_t(i)\\&=&
\arg\max_{h\in\mathcal{H}^1}\Big{(}1-2\sum_{h(\bm{x}_i)\neq \tilde{r}_{t,i}}w_t(i)\Big{)}=
\arg\min_{h\in\mathcal{H}^1}Pr_{i\sim w_t}[h(\bm{x}_i)\neq \tilde{r}_{t,i}],
\end{eqnarray*}
which means that the maximization subproblem is equivalent to minimizing the weighted classification error. Thus, in this fashion,  $\arg\max_{h\in\mathcal{H}^1}\langle h,\bm{r}_t\rangle$ can be reduced to a sequence of classification problems.  The resulting practical algorithm is described in Algorithm \ref{FWC} as FWBoost\_C.

Alternatively, for general base hypothesis space $\mathcal{H}$ taking values in $\mathbb{R}$, if $\mathcal{H}$ is closed under scalar multiplication (which is true for commonly used base hypotheses like regression trees, splines or stumps), we can instead solve an alternative problem $h^*=\arg\min_{h\in\mathcal{H}}\|h-\bm{r}_t\|_2$ by directly minimizing the Euclidean distance and then let $h_t=Ch^*/\|h^*\|_\infty$. Now finding such an $sÕ$ is itself a least-squares regression problem with real-valued negative gradient $\bm{r}_t$, as described in Algorithm \ref{FWR} as FWBoost\_R. 

\begin{minipage}{.47\linewidth}
\begin{algorithm}[H]\label{FWC}
\caption{FWBoost\_C}
\SetKwData{Left}{left}\SetKwData{This}{this}\SetKwData{Up}{up}
\SetKwFunction{Union}{Union}\SetKwFunction{FindCompress}{FindCompress}
\SetKwInOut{Input}{input}\SetKwInOut{Output}{output}
 \SetAlgoLined
... {\it{Replace line (a) in Algorithm \ref{FWB} with}} \\
$\tilde{r}_{t,i}=\mathrm{sign}(r_{t,i})$, $w_t(i)=|r_{t,i}|/\|\bm{r}_t\|_1$\\
Train base classifier $h^* \in \mathcal{H}$ using  $w_t$ and  $\tilde{r}_{t,i} $\\
$h_t=Ch^*$
\end{algorithm}
\end{minipage} \hfill
\begin{minipage}{.5\linewidth}
\begin{algorithm}[H]\label{FWR}
\caption{FWBoost\_R}
\SetKwData{Left}{left}\SetKwData{This}{this}\SetKwData{Up}{up}
\SetKwFunction{Union}{Union}\SetKwFunction{FindCompress}{FindCompress}
\SetKwInOut{Input}{input}\SetKwInOut{Output}{output}
 \SetAlgoLined
... {\it{Replace line (a) in Algorithm \ref{FWB} with}} \\
Fit base hypothesis $h^* \in \mathcal{H}$ to residuals $r_{t,i}$ by least squares regression on training set $\{(\bm{x}_i, r_{t,i})\}_{i=1}^m$:
$h_t=Ch^*/\|h^*\|_\infty$
\end{algorithm}
\end{minipage} 

This is computationally attractive for dealing with non-parametric learning problems with large or infinite number of base hypotheses and general loss functions.  First, the computational effort in Frank-Wolfe Boosting  is exactly the same as in AdaBoost or gradient boosting to refit the residuals with no other complex optimization procedure needed, which ensures the efficiency of this method even for large scale datasets. Second, it does not need to identify whether a newly added base hypothesis is the same as one  already included from previous steps (e.g. as done in BLasso \cite{zhao2004boosted}), which is awkward in practice especially for a large and complex base hypothesis space. Therefore it is free from the need to  search through previously added base hypotheses. Thus  the computation complexity at each iteration will not increase. 

Note that if we solve the subproblem $(a)$ exactly and if the FW direction $s_t-F_{t-1}$ turns out to be a non-decreasing direction, then we can conclude that we are already at the global minimum. More details are provided in the supplementary material.

Guelat and Marcotte  \cite{guelat1986some} developed Frank-Wolfe with away-steps to directly solve the sparsity problem. Here we provide the away-step variant of Frank-Wolfe boosting method, which directly solves the $l_1$ constrained regression problem while maintaining the sparsity of the base hypothesis by away-steps that potentially  remove an old base hypothesis with bad performance. The detailed algorithm is written as the following

\begin{algorithm}[H]\label{algorithm:away}
\caption{Frank-Wolfe Gradient Boosting with awaystep}
\SetKwData{Left}{left}\SetKwData{This}{this}\SetKwData{Up}{up}
\SetKwFunction{Union}{Union}\SetKwFunction{FindCompress}{FindCompress}
\SetKwInOut{Input}{input}\SetKwInOut{Output}{output}
 \SetAlgoLined

 \Input{$m$ examples $(x_1,y_1)\dots(x_m,y_m)$ and constant $C$}
 Set $\mathcal{I}_1=\emptyset$
  \BlankLine

 \For{$t=1,2,3,\dots T$ }{
$r_{t,i}=-\big[\frac{\partial L(y_i,F(x_i))}{\partial F(x_i)}\big]_{F(x)=F_{t-1}(x)},~~i=1,\dots,m$\\
Solve the subproblem $h_t=C\cdot\arg\min_{h\in \mathcal{H}^1}\langle h,\bm{r}_t\rangle$, define FW direction $d_t=h_t-F_{t-1}$\\

Solve the subproblem $h_{min}=\arg\min_{h\in \mathcal{I}_t} \langle h,\bm{r}_t\rangle$, define away direction $d^A_t=F_{t-1}-Ch_{min}$\\

\eIf {$d_t^{FW}\cdot \bm{r}_t>d_t^A\cdot \bm{r}_t$ $||~t==1$}
{
$\gamma_t^{FW} =\arg\min_{\gamma\in[0,1]}\sum_{i=1}^{m}L\big{(}y_i,F_{t-1}(x_i)+\gamma d_t^{FW}(x_i)\big{)}$\\
Perform the FW step $\alpha_k\leftarrow (1-\gamma_t^{FW} )\alpha_k$ for $h_k\in \mathcal{I}_t$ and $\alpha_t^{FW}=\gamma_t^{FW} C$\\
Let $h_t^{FW}=h_t/C$ and update $\mathcal{I}_t$ by $\mathcal{I}_{t+1}=\mathcal{I}_{t}\cup \{h_t^{FW}\}$\\
$F_t=F_{t-1}+\gamma_t^{FW}d_t^{FW}$
}
{
$\gamma_t^A =\big{\{}\alpha_{min}/(C-\alpha_{min}),\arg\min_{\gamma\in[0,1]}\sum_{i=1}^{m}L\big{(}y_i,F_{t-1}(x_i)+\gamma d_t^{A}(x_i)\big{)}\big{\}}$\\
Perform the away step $\alpha_k\leftarrow (1+\gamma_t^{A} )\alpha_k$ for $h_k\neq h_{min}$ and $\alpha_t^{FW}=(1+\gamma_t^{FW}) \alpha_t^{FW}-\gamma_t^{FW}C$\\
\textbf{if} $\gamma_t^A=\alpha_{min}/(C-\alpha_{min})$  \textbf{then} $\mathcal{I}_{t+1}=\mathcal{I}_t\backslash \{h_{min}\}$\\
$F_t=F_{t-1}+\gamma_t^{A}d_t^{A}$
}
}

\Output{$F_T(x)$}
\end{algorithm}

\subsection{Theoretical Analysis}
In this section, we provide theoretical guarantees for empirical risk and generalization of functional Frank-Wolfe boosting for general loss functions. \subsubsection{Rademacher Complexity  Bounds}
We use Rademacher complexity as a standard tool to capture the richness of a family of functions in the analysis of voting methods \cite{koltchinskii2002empirical}.

Let $\mathcal{G}$ be a family of functions mapping from $\mathcal{X}$ to $[a,b]$, and let $S=(\bm{x}_1,...,\bm{x}_m)$ a fixed  i.i.d. sample set with $m$ elements in $\mathcal{X}$. The empirical Rademacher complexity  of  $\mathcal{G}$ with respect to $S$ is defined as
\begin{equation*}
\hat{\mathcal{R}}_{\mathcal{S}}(\mathcal{G})=E_{\sigma}\Big{[}\sup_{g\in\mathcal{G}}\frac{1}{m}\sum_{i=1}^{m}\sigma_ig(x_i)\Big{]},
\end{equation*}
where the $\sigma_i$'s are independent uniform random variables taking values in $\{-1,+1\}$.  We also define the risk and empirical risk of $h$ as $\mathbb{E}[l_h] := \mathbb{E}_{(\bm{x},y)}[l(h(\bm{x}),y)]$  and $
\hat{\mathbb{E}}[l_h] := \frac{1}{m} \sum_{i=1}^m l( h(\bm{x}_i),y_i)
$, respectively. By Talagrand's lemma and general Rademacher complexity uniform-convergence bounds \cite{mohri2012foundations}, we have the following Rademacher complexity bound for general loss functions. The bound is in terms of empirical risk, the sample size and the complexity of $\mathcal{F}$.

\begin{theorem} \label{gb} Let $\mathcal{F}$ be a set of real-valued functions. Assume the loss function $l$ is $L_l$-Lipschitz continuous with respect to its first argument and that $l(y, y') \le M$,  $\forall y, y' \in \mathcal{Y}$. For any $\delta > 0$ and with probability at least $1- \delta$ over a sample $S$ of size $m$, the following inequalities hold for all $f \in \mathcal{F}$:
$$
\mathbb{E}\left[ l_f\right] \le \hat{\mathbb{E}}[l_f]+ 2L_l\hat{\mathcal{R}}_S(\mathcal{F})+3M\sqrt{\frac{\log \frac{2}{\delta}}{2m}}.
$$
\end{theorem}
We next show an upper bound for the Rademacher complexity of a regularized function space $\mathcal{F_T^C}$ after $T$-step boosting \cite{koltchinskii2002empirical}. Namely that complexity, as measured by Rademacher, is the same for the combined class as for the base class times the constant $C$. The proof is left in the supplement.

\begin{theorem}\label{aa} For any sample $S$ of size $m$, 
$\hat{\mathcal{R}}_S(\mathcal{F}_T^C) \le C\hat{\mathcal{R}}_S(\mathcal{H}^1).$ Further if the assumption that  $\mathcal{H}$ is 
closed  under  scalar multiplication holds, $\hat{\mathcal{R}}_S(\mathcal{F}_T^C) = C\hat{\mathcal{R}}_S(\mathcal{H}^1).$
\end{theorem}

Based on Theorems \ref{gb} and \ref{aa}, we can see that the risk bound is depending on the $l_1$ regularization constant $C$ and is independent of the number of boosting iterations. Unlike vanilla gradient boosting ( even with shrinkage and subsampling to prevent overfitting), the  generalization error will not increase even if the algorithm is boosted forever. Specifically, the Rademacher complexity bounds for  regression with $l_p$ loss are given in the supplementary material.

\subsection{Bounds on Empirical Risk}
In this subsection we analyze the convergence rate of Frank-Wolfe boosting on the training set.
During the Frank-Wolfe subproblem, we linearize the objective function to get a descent direction.
To analyze the training error of our algorithm, we first study the second derivative of the loss function in the feasible region. If $|\frac{\partial^2 l}{\partial a^2}(a,y)|\le C_l$ for $a\in \mathcal{Y}$,
then by Taylor's theorem $l (a+\lambda(b-a),y)\le l(a,y)+\lambda\langle b-a,\frac{\partial l}{\partial a}\rangle+\frac{\lambda^2}{2}(b-a)^2C_l$, which gives us enough evaluation of the distance between $l$ and its linearization. Noticing that the boundedness of the second derivative is conserved under summation,  this property passes to the empirical risk $L(F(\bm{x}),\bm{y})=\frac{1}{m}\sum_{i=1}^ml(F(\bm{x}_i),y_i)$ for $F(\bm{x})\in\mathcal{Y}$.
This leads us to the following condition which we require for our convergence analysis.

\textbf{Assumption 1} (Smooth Empirical Risk) For empirical risk function $L(F,\bm{y})$, there exists a constant $C_{l,\mathcal{F}}$ that depends on loss $l$ and predictor space $\mathcal{F}$, such that for any $f,g\in \mathcal{F}$, we have
$$L (f+\lambda(g-f),\bm{y})\le L(f,\bm{y})+\lambda\langle g-f,\frac{\partial L}{\partial F}\Big{|}_{F=f}\rangle+\frac{\lambda^2}{2}C_{l,\mathcal{F}}.$$

This assumption is valid for most commonly used losses for bounded $\mathcal{Y}$. First, notice that in our constrained boosting setting, for any predictor $f\in\mathcal{F}^C$, based on H$\ddot{o}$lder's inequality, we have
$|f|=|\bm{\alpha}\cdot h(\bm{x})|\le\|\bm{\alpha}\|_1\|h(\bm{x})\|_\infty\le C$.
This specifies the range in which this Assumption 1 needs to hold.
For $l_p$ loss $l(a,y)=(a-y)^p/p, (p\ge 2)$, $\frac{\partial^2 l}{\partial a^2}(a,y)=(p-1)(a-y)^{p-2}\le (p-1)(2C)^{p-2}$ and $(a-y)^2\le 4C^2$, so $C_{l,\mathcal{F}^C}$ can be chosen as $(p-1)(2C)^p$.
For exponential loss $l(a,y)=\exp(-ay)$, $y\in\{1,-1\}$, $|\frac{\partial^2 l}{\partial a^2}(a,y)|=|y^2\exp(-ay)|\le e^C$, so $C_{l,\mathcal{F}^C}$
can be chosen as $4C^2e^C$.
For logistic loss $l(a,y)=\log(1+\exp(-ay))$, $y\in\{1,-1\}$, $|\frac{\partial^2 l}{\partial a^2}(a,y)|=y(1+\exp(-ay))^{-1}(1+\exp(ay))^{-1}|\le\frac{1}{4}$,  so $C_{l,\mathcal{F}^C}$ can be chosen as $C^2$.

\begin{theorem}
Under Assumption 1, for convex loss $l$, suppose we could exactly solve the linearized subproblem $h_t=\arg\max_{s\in\mathcal{H}^1}\langle s,\mathbf{r}_t\rangle$ at each iteration, and the step size $\gamma_t$ is as $\gamma_t=2/(t+2)$, then the training error for Frank-Wolfe boosting algorithm is bounded:
$L(F_t,y)-L(F^*,y)\le \frac{C^*}{2+t},$
where $F^*$ is the optimal solution to \eqref{Fobj} and the constant $C^*=\max\{\frac{C_{l,\mathcal{F}^C}}{2},3L(0,\bm{y})/4\}$ depends on the predictor space $\mathcal{F}^C$, loss function $l$ and initial empirical loss $L(0,\bm{y})$.
\end{theorem}

In the learning setting, the subproblem $h_t=\arg\max_{s\in\mathcal{D}}\langle s,\mathbf{r}_t\rangle$ is  solved by fitting the residual by least-square regression or finding the best classifier. However the  base learning algorithm sometimes only finds an approximate maximizer rather than the global maximizer.
Nevertheless, if we assume the $\arg\max$ problem is solved not too inaccurately at each step, we have the following error bound. The detailed proof of all the theorems in this section are left in the supplementary materials.  
\begin{theorem}
Under the same assumption and notation, for a positive constant $\delta$, if the linearized subproblem is solved with tolerance $\delta\lambda_tC_{l,\mathcal{F}}$,  that is
$$\langle h_t,\mathbf{r}_t\rangle\ge\max_{s\in\mathcal{D}}\langle s,\mathbf{r}_t\rangle-\delta\lambda_tC_{l,\mathcal{F}},$$
then we have the following empirical risk bound
$$L(F_t,y)-L(F^*,y)\le \frac{C^*}{2+t}(1+2\delta).$$

\end{theorem}

\section{Frank-Wolfe Boosting for classification}\label{sec:Class}
AdaBoost is one of the most popular and important methods for classification problem where $y \in \{-1, +1\}$.
It is well known that AdaBoost can be derived from functional gradient decent method using  exponential loss \cite{breiman1998arcing,schapire2012boosting}. The margin theory predicts  AdaBoost's resistance to overfitting provided that large margins can be achieved and the base hypotheses are not too complex. Yet overfit can certainly happen when the base hypothesis is not able to achieve large margins; or when the noise is overwhelming; or when the space of base hypotheses is too complex. Following the same path, we apply the Frank-Wolfe boosting algorithm for exponential loss with $l_1$ constrained regularization and derive a classification method with a similar reweightening procedure as in AdaBoost. Yet it comes with a tunable parameter $C$ to balance the empirical risk and generalization.

Specifically, define the exponential loss function $l(F,y)=e^{-yF(\bm{x})}$ and a base classifier space $\mathcal{H}$. Following the same logic as in Algorithm \ref{FWC}, we first calculate the negative gradient at each round as $r_{t,i}=\frac{y_ie^{-y_iF_{t-1}(\bm{x}_i)}}{m}$.  On round $t$, the goal is to find $h_t=\arg\max_{h \in \mathcal{H}}\langle h_t,\bm{r}_t\rangle$ which is proportional to $\sum_{i=1}^mD_t(i)y_ih_t(\bm{x}_i)$ and is equivalent to minimizing the weighted classification error with weight vector $D_t(i)=\frac{e^{-y_iF_{t-1}(\bm{x}_i)}}{\sum e^{-y_iF_{t-1}(\bm{x}_i)}}$. After $h_t$ has been chosen, that stepsize $\gamma_t$ can be selected using the methods already discussed, such as  line search. After setting $\gamma_t$, we are ready to update the weights: 
$$
D_{t+1}(i)\propto e^{-y_iF_{t}(\bm{x}_i)}=e^{-y_i(1-\gamma_t)F_{t-1}(\bm{x}_i)}e^{-C\gamma_ty_ih_t(\bm{x}_i)}\propto D_t^{1-\gamma_t}(i)e^{-C\gamma_ty_ih_t(\bm{x}_i)}.
$$

With exponential loss, the generic FWBoost algorithm  for classification can be rewritten in a reweightening procedure like AdaBoost in Algorithm \ref{EFWB}, where $\alpha_{[1:t]}$ denotes the vector $[\alpha_1,...,\alpha_t]$. AdaBoost.FW is computationally efficient with large or infinite hypothesis space while a similar non-practical counterpart suitable for a small number of base hypotheses is mentioned in \cite{slides}.

\begin{algorithm}[H]
\caption{AdaBoost.FW: a  functional Frank-Wolfe algorithm for classification} \label{EFWB}
\SetKwData{Left}{left}\SetKwData{This}{this}\SetKwData{Up}{up}
\SetKwFunction{Union}{Union}\SetKwFunction{FindCompress}{FindCompress}
\SetKwInOut{Input}{input}\SetKwInOut{Output}{output}
 \SetAlgoLined

 \Input{$m$ examples $(\bm{x}_1,y_1),(\bm{x}_2,y_2),\dots(\bm{x}_m,y_m)$ and constant $C>0$}
  \BlankLine
Initialize:  $D_1(i)=1/m$ for $i=1,...,m.$\\
 \For{$t=1$ to $T$}{
 Train weak hypothesis using distribution  $D_t$ and get $h_t \in \mathcal{H}$\\
Choose $\gamma_t \in [0,1]$, for $i,1,...,m$ with $Z_t$ a normalization factor: $D_{t+1}(i)=\frac{D_t^{1-\gamma_t}(i)}{Z_t} \times e^{-\gamma_tCy_ih_t(\bm{x}_i)}$\\

$\alpha_t$:=$\gamma_tC$ and $\alpha_{[1:t-1]}\leftarrow(1-\gamma_t)\alpha_{[1:t-1]}$ \\
}
\Output{$F_T(\bm{x})=\text{sign}\big(\sum_{t=1}^T\alpha_th_t(\bm{x})\big)$}
\end{algorithm}

\section{Experimental Results}\label{sec:exp}
In this section, we evaluate the proposed method on UCI datasets \cite{Lichman:2013}. In order to randomize the experiments, in each run of experiments we  select 50\% data as the training examples. The remaining 50\% data is used as test set. Furthermore, in each run, the parameters of each method are tuned by 5-fold cross validation on the training set. This random split was repeated 20 times and the results are averaged over 20 runs.
\paragraph{Least-squares Regression}
To demonstrate the effectiveness of our proposed method in avoiding overfitting while achieving reasonably good training error, we compare FWBoost\_C and FWBoost\_R with existing state-of-the-art boosting algorithms (with regularization), including vanilla Gradient Boosting \cite{friedman2001greedy},  Gradient Boosting with shrinkage \cite{hastie2009elements}, Gradient Boosting with early stopping \cite{hastie2009elements}, Gradient Boosting with subsampling \cite{friedman2002stochastic} and BLasso \cite{zhao2004boosted} for least-squares regression problems. 

\begin{figure}[htp!]
    \centering
    \begin{tabular}{ccc}
       \includegraphics[width=0.3\textwidth]{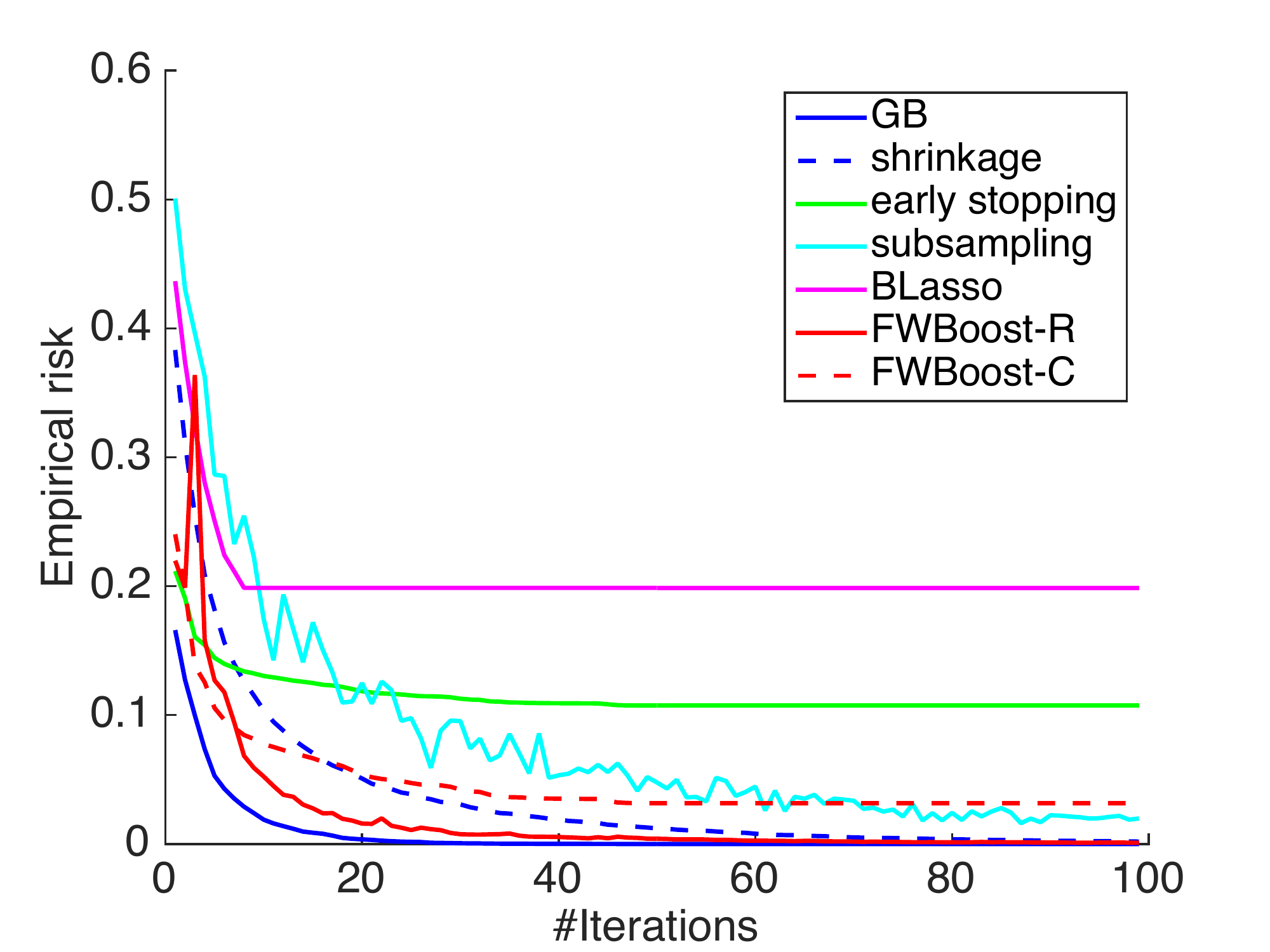}
 \includegraphics[width=0.3\textwidth]{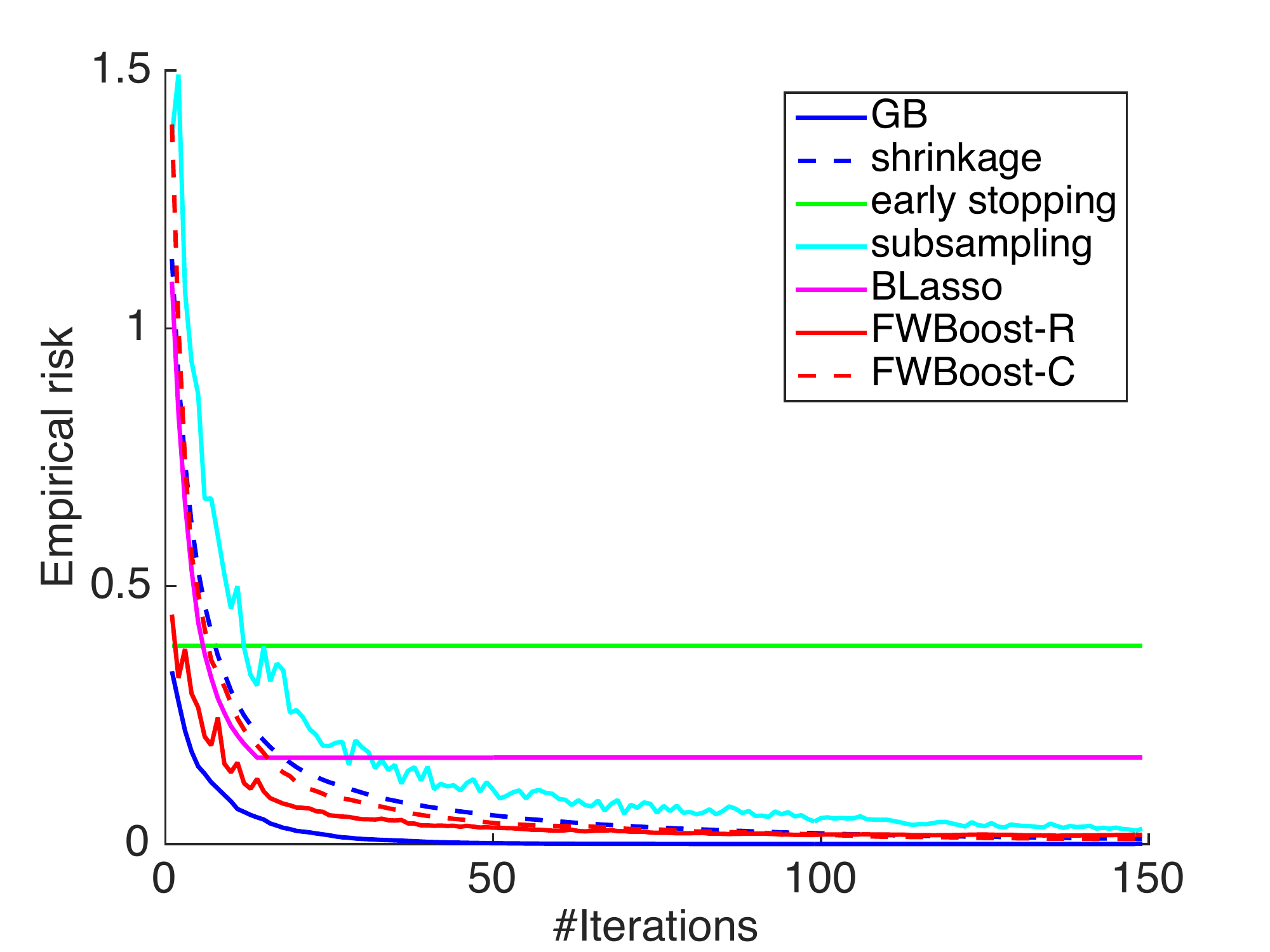} 
 \includegraphics[width=0.3\textwidth]{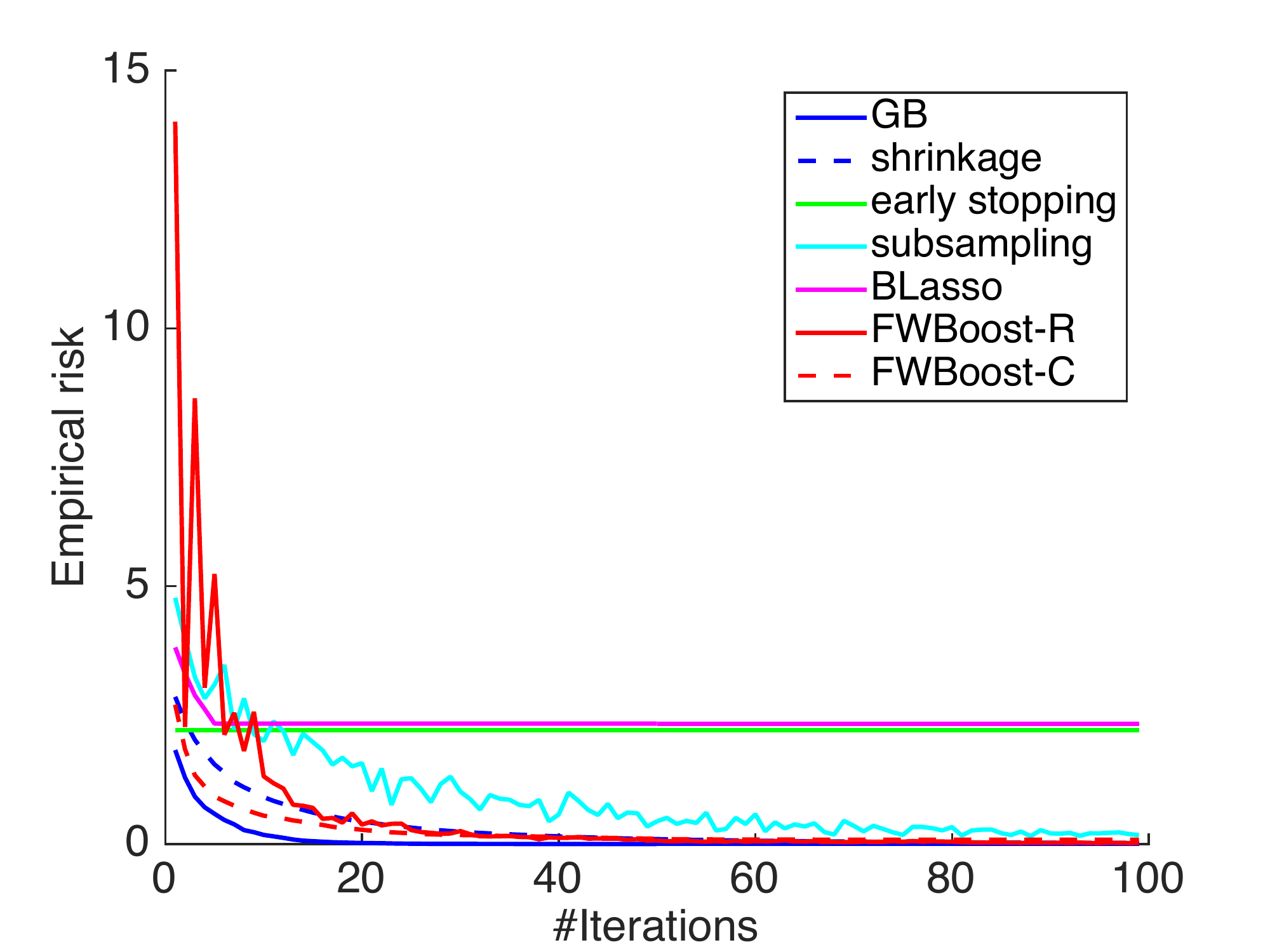} \\
 \subfigure[Auto MPG]{
        \includegraphics[width=0.3\textwidth]{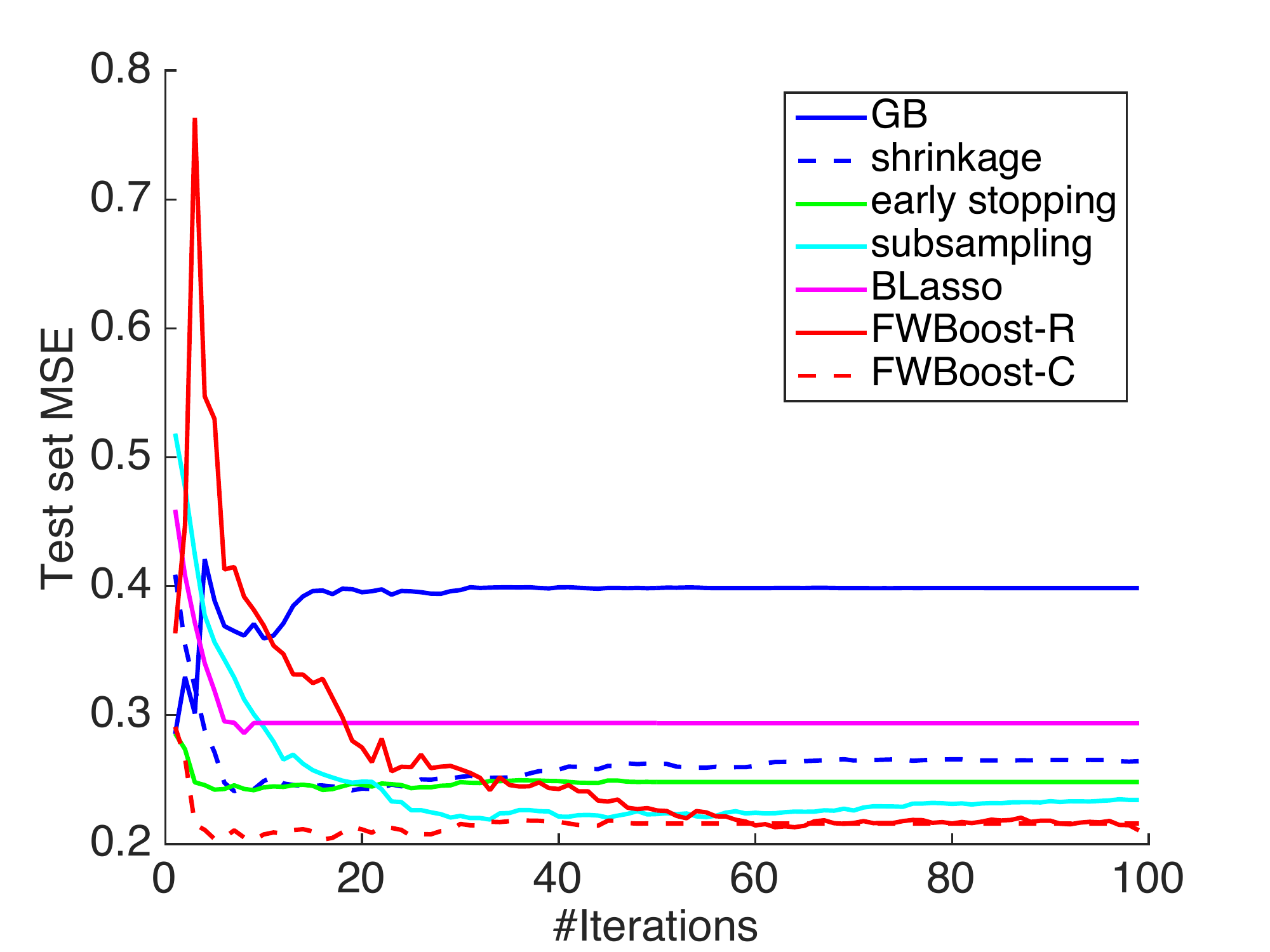}}
 \subfigure[Housing]{ \includegraphics[width=0.3\textwidth]{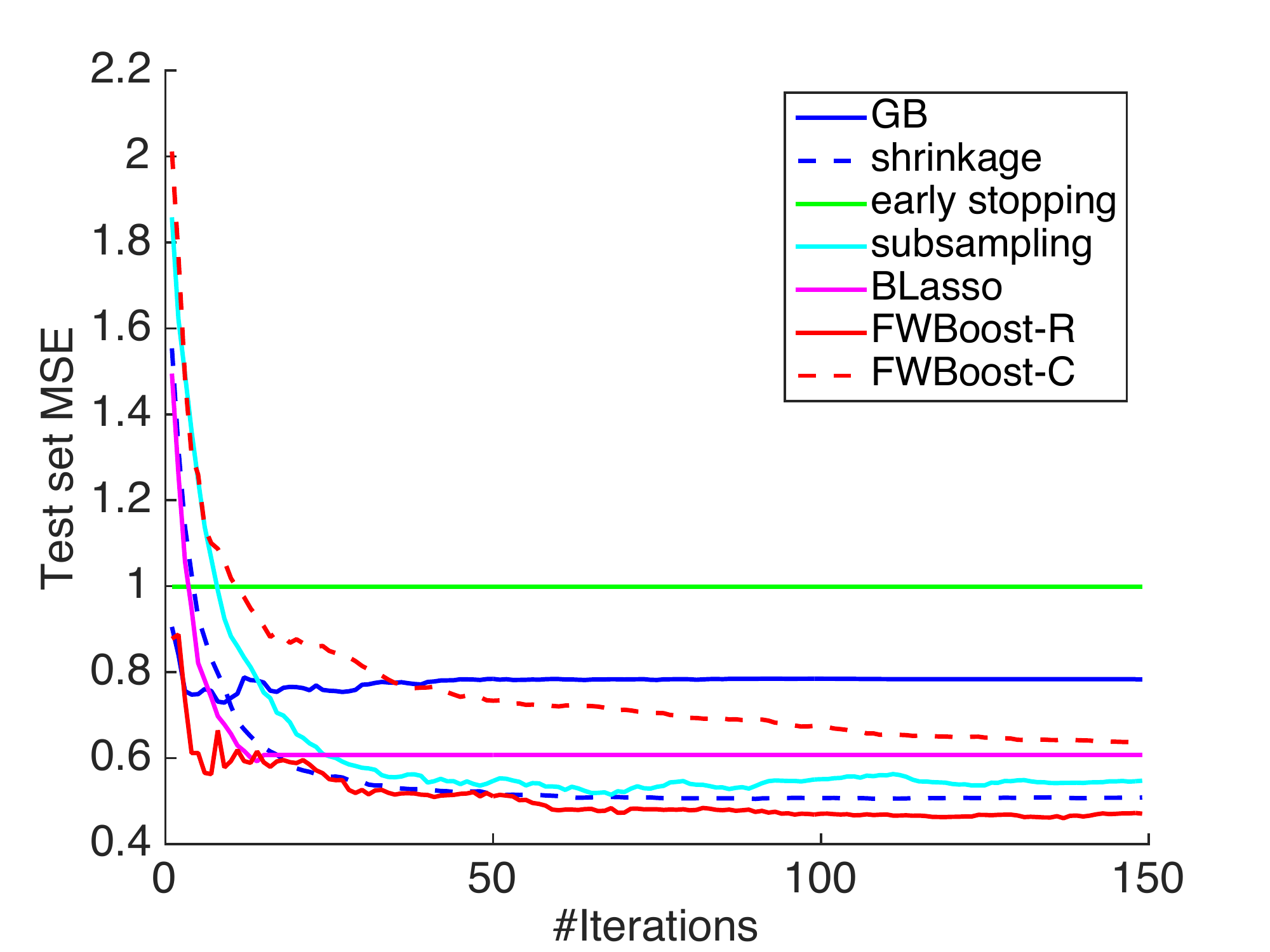} }
 \subfigure[Concrete Slump]{ \includegraphics[width=0.3\textwidth]{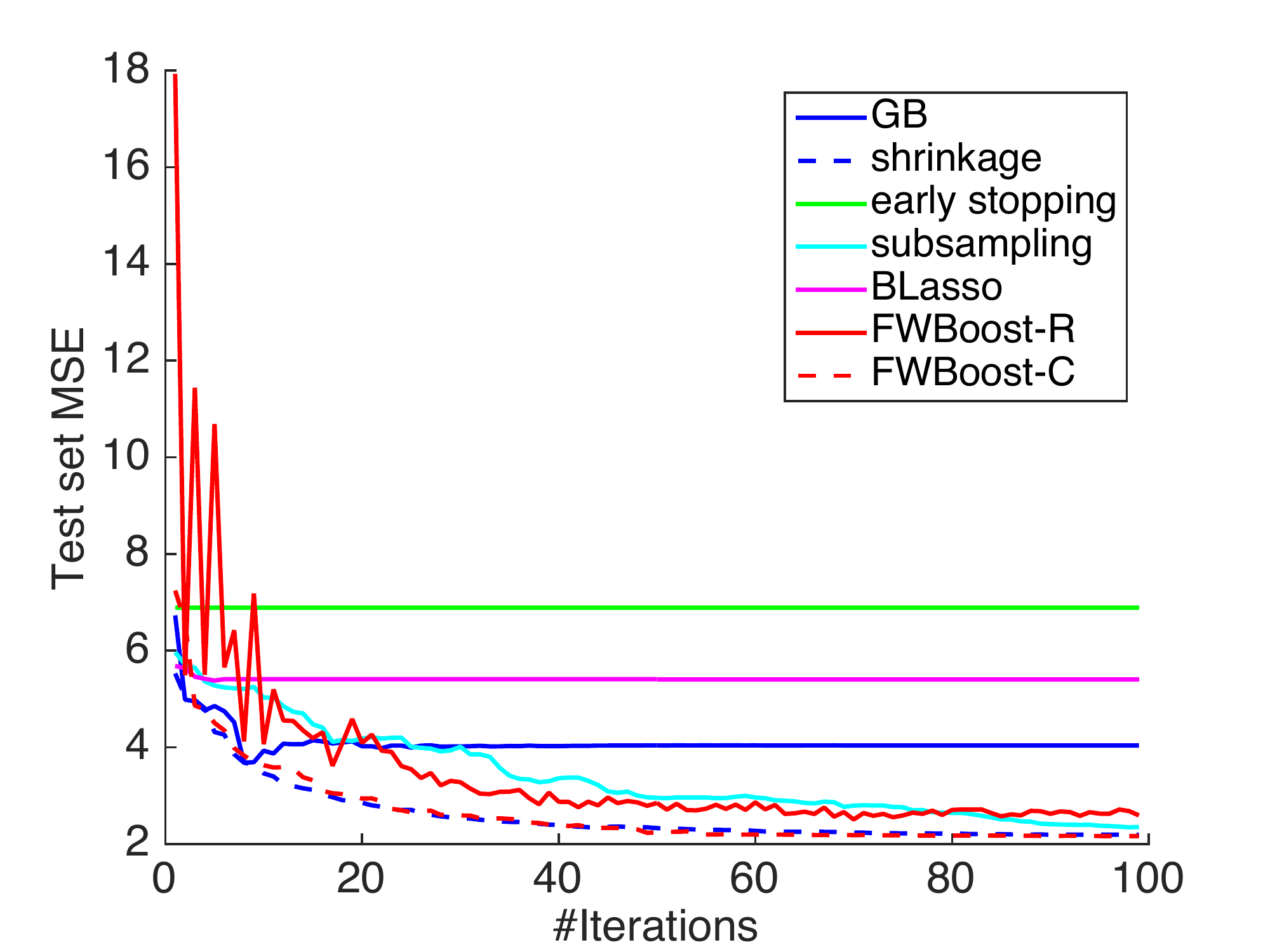} }
    \end{tabular}
    \caption{Comparison of boosting methods on UCI datasets. The x-axis is the number of boosting iterations. The first row is the averaged empirical risk and the second row is the MSE on test set. \label{f1}}
\end{figure}

The experimental results are shown in Fig. \ref{f1}.   The first row is the empirical risk averaged  over 20 runs and the second row is the averaged MSE on test set. For early stopping and BLasso, if the algorithm stops  before reaching the maximum iteration, the  MSE are set to their last rounds' values.  It is demonstrated in Fig. \ref{f1}. that FWBoost algorithms reduce the empirical risk considerably fast. In all the examples where gradient boosting overfits after early iterations, FWBoost algorithms achieve the minimal averaged test risk compared to other regularization methods. It is consistent with the theoretical analysis that no degradation in performance of FWBoost is observed with larger rounds of boosting. In addition, it is difficult to observe backward steps of BLasso algorithm because it usually stops at early iterations. It behaves similarly to early stopping method.

\paragraph{Binary Classification}
In order to demonstrate the effectiveness of AdaBoost.FW, we compare with AdaBoost on UCI data sets that cause AdaBoost to overfit. The first example is the heart-disease data set as presented in \cite{schapire2012boosting} as an example of overfitting. Decision stumps are used as weak learners. While reducing the training error,  the test error of AdaBoost method may go up after several iterations.
On the other hand, AdaBoost.FW does not suffer from overfitting and achieves smaller test error.

\begin{figure}[htp!]
    \centering
    \begin{tabular}{ccc}
       \includegraphics[width=0.3\textwidth]{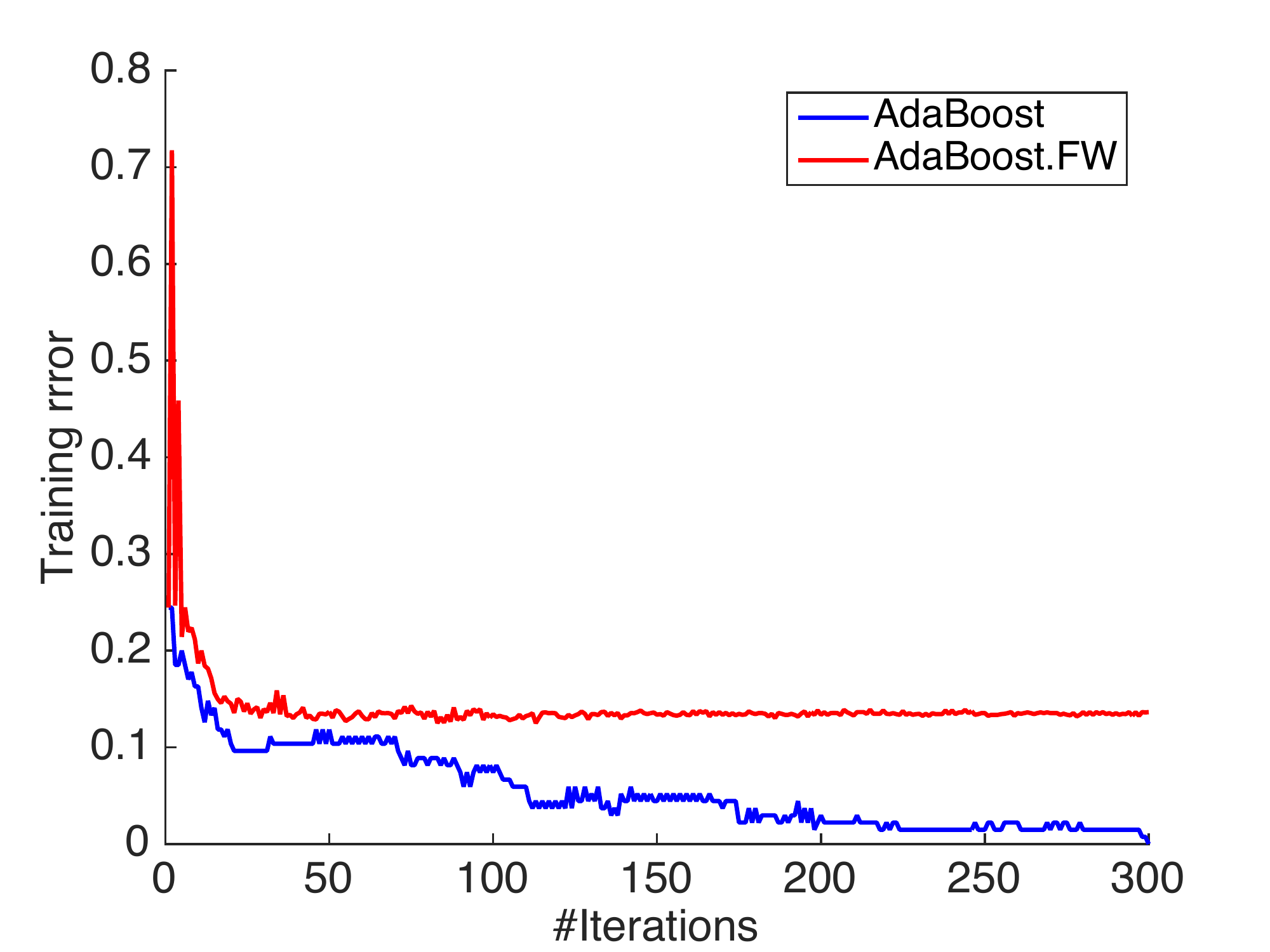}
 \includegraphics[width=0.3\textwidth]{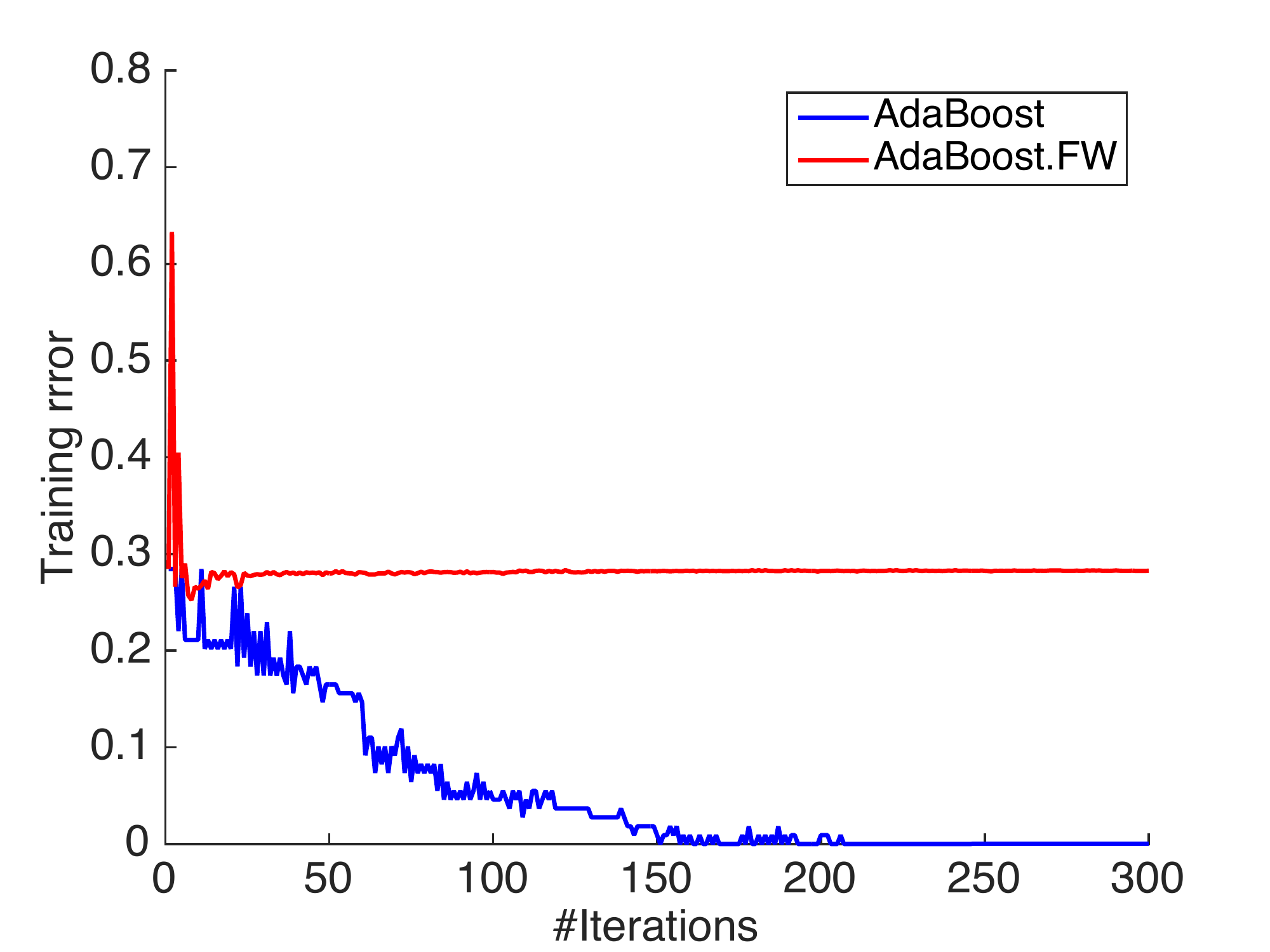} 
 \includegraphics[width=0.3\textwidth]{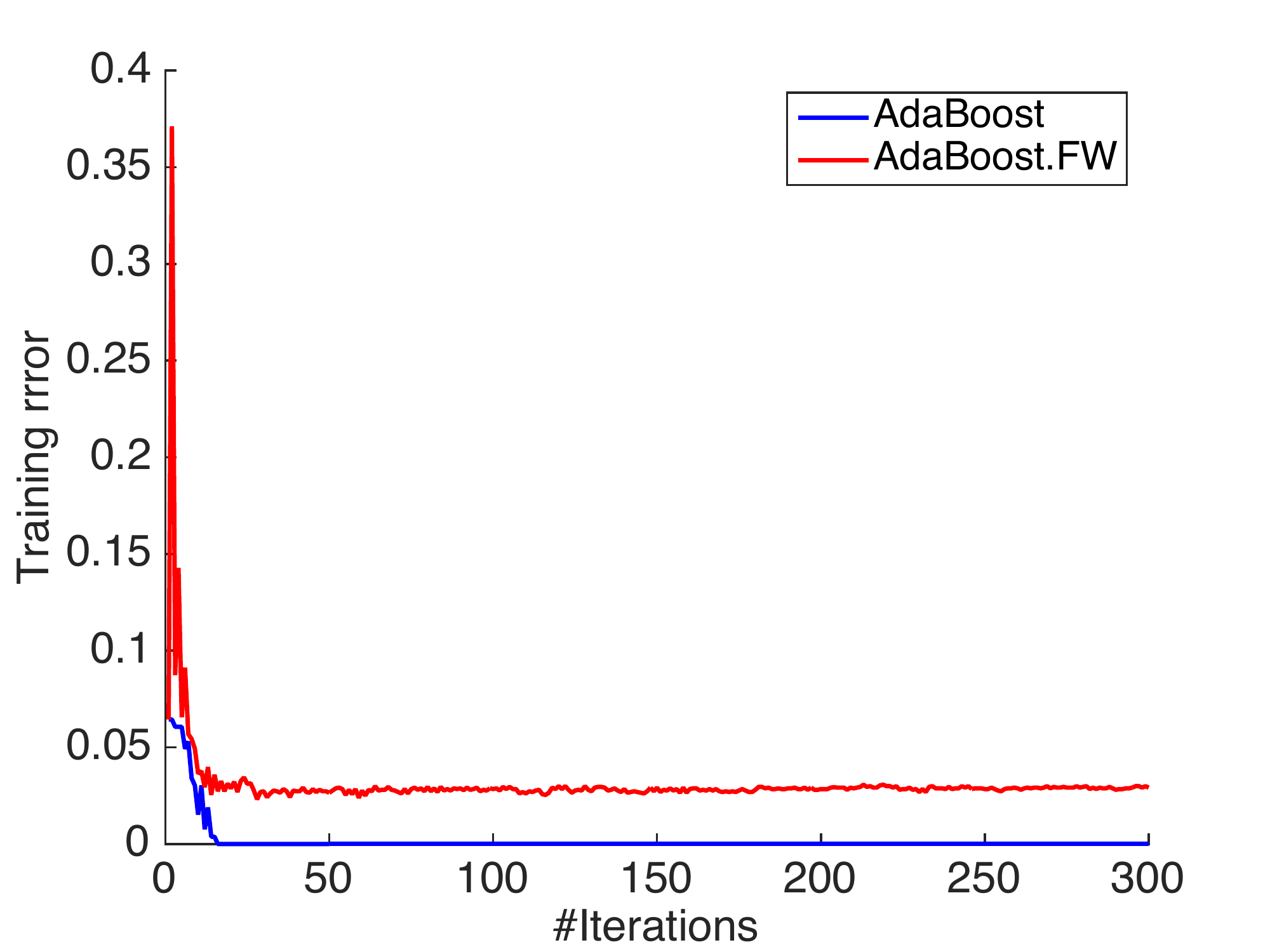} \\
 \subfigure[Heart disease]{
        \includegraphics[width=0.3\textwidth]{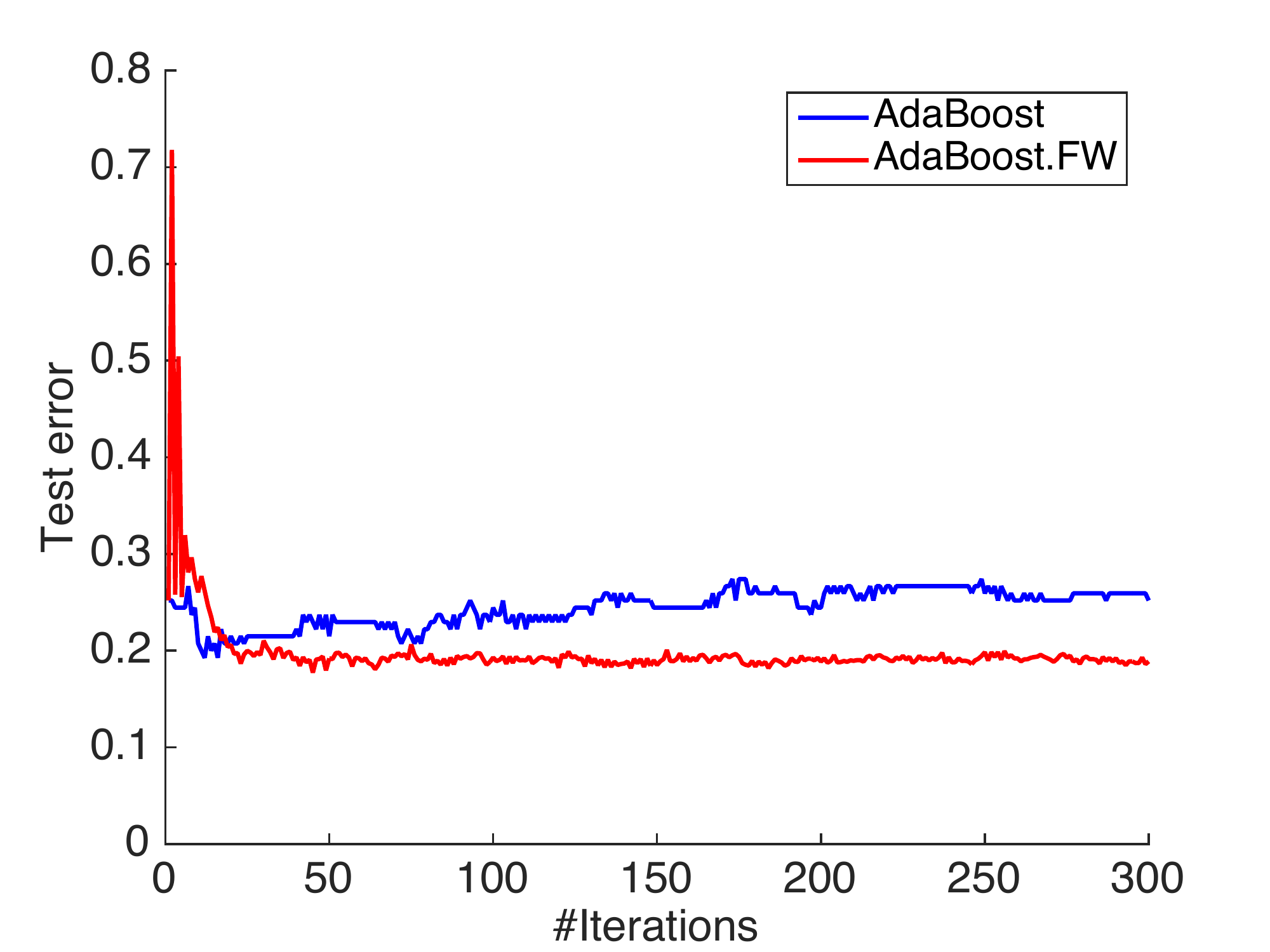}}
 \subfigure[Planning relax]{ \includegraphics[width=0.3\textwidth]{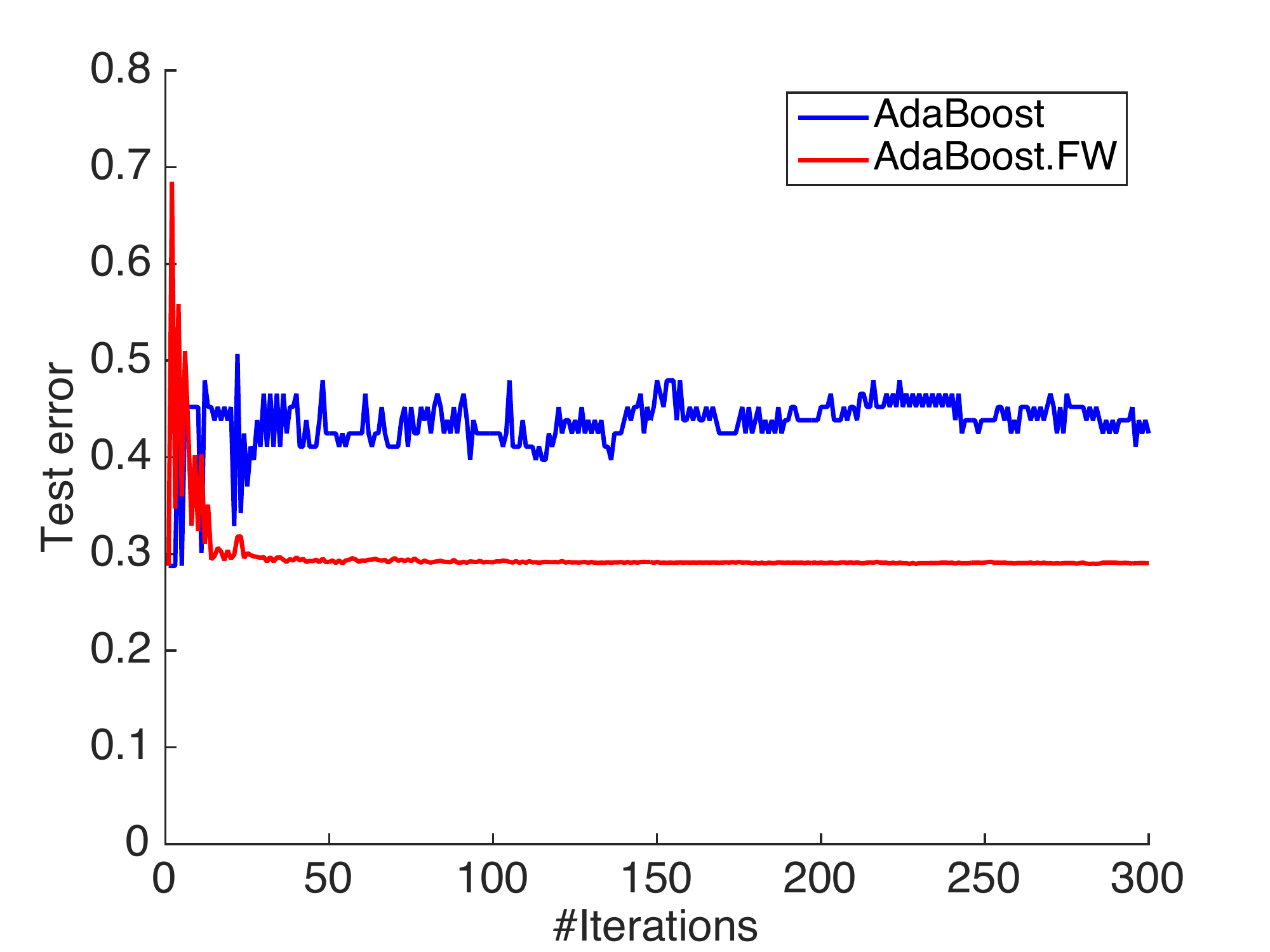} }
 \subfigure[Wholesale customers]{ \includegraphics[width=0.3\textwidth]{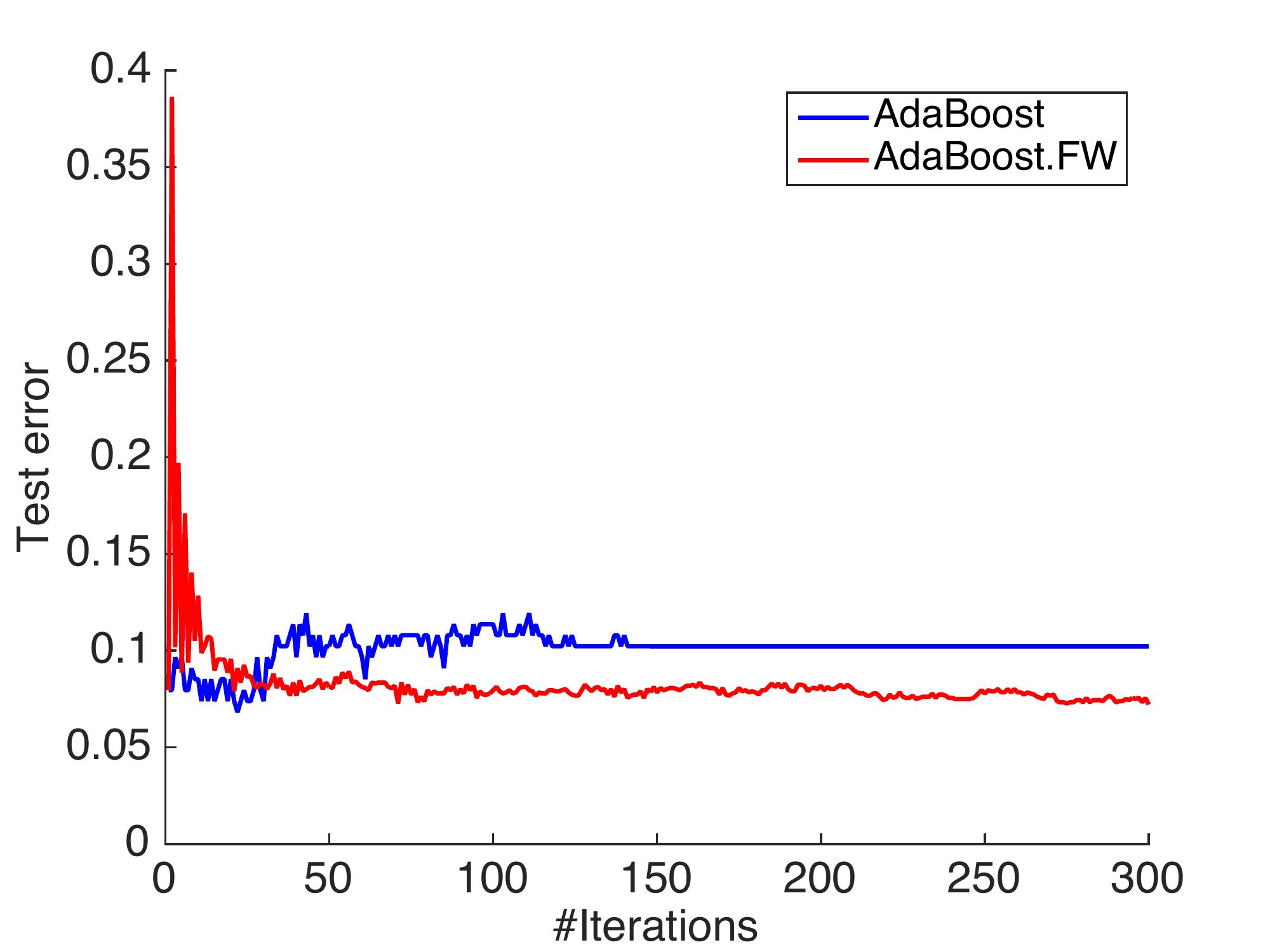} }
    \end{tabular}
    \caption{Comparison of boosting methods on UCI datasets. The x-axis is the number of boosting iterations. The first (second) row is the averaged training (test) error over 20 runs. \label{f2}}
\end{figure}

\section{Conclusion}
In this paper, we considered the $l_1$ regularized  loss function minimization for ensemble learning to avoid overfitting. We propose a novel Frank-Wolfe type boosting algorithm applied to general  loss functions and analyze empirical risk and generalization.  By using exponential loss for binary classification, the FWBoost algorithm can be rewritten as an AdaBoost-like reweightening algorithm. Furthermore, it is computationally efficient  since the computational effort in FWBoost is exactly the same as in AdaBoost or gradient boosting to refit the residuals. We further deploy an important variant of Frank-Wolfe algorithms with away-steps to improve sparsity of boosting algorithms.

\bibliography{refer}

\begin{thebibliography}{10}

\bibitem{breiman1998arcing}
Leo Breiman.
\newblock Prediction games and arcing algorithms.
\newblock {\em Neural computation}, 11(7):1493--1517, 1999.

\bibitem{friedman2001greedy}
Jerome~H Friedman.
\newblock Greedy function approximation: a gradient boosting machine.
\newblock {\em Annals of statistics}, pages 1189--1232, 2001.

\bibitem{mason1999functional}
Llew Mason, Jonathan Baxter, Peter~L Bartlett, and Marcus Frean.
\newblock Functional gradient techniques for combining hypotheses.
\newblock {\em Advances in Neural Information Processing Systems}, pages
  221--246, 1999.

\bibitem{friedman2000additive}
Jerome Friedman, Trevor Hastie, Robert Tibshirani, et~al.
\newblock Additive logistic regression: a statistical view of boosting.
\newblock {\em The annals of statistics}, 28(2):337--407, 2000.

\bibitem{freund1997decision}
Yoav Freund and Robert~E Schapire.
\newblock A decision-theoretic generalization of on-line learning and an
  application to boosting.
\newblock {\em Journal of computer and system sciences}, 55(1):119--139, 1997.

\bibitem{schapire2003boosting}
Robert~E Schapire.
\newblock The boosting approach to machine learning: An overview.
\newblock In {\em Nonlinear estimation and classification}, pages 149--171.
  Springer, 2003.

\bibitem{schapire1998boosting}
Robert~E Schapire, Yoav Freund, Peter Bartlett, and Wee~Sun Lee.
\newblock Boosting the margin: A new explanation for the effectiveness of
  voting methods.
\newblock {\em Annals of statistics}, pages 1651--1686, 1998.

\bibitem{grove1998boosting}
Adam~J Grove and Dale Schuurmans.
\newblock Boosting in the limit: Maximizing the margin of learned ensembles.
\newblock In {\em AAAI/IAAI}, pages 692--699, 1998.

\bibitem{buhlmann2002consistency}
Peter~Lukas Buhlmann.
\newblock Consistency for l2boosting and matching pursuit with trees and
  tree-type basis functions.
\newblock In {\em Book}, 2002.

\bibitem{zhang2005boosting}
Tong Zhang and Bin Yu.
\newblock Boosting with early stopping: convergence and consistency.
\newblock {\em Annals of Statistics}, pages 1538--1579, 2005.

\bibitem{friedman2002stochastic}
Jerome~H Friedman.
\newblock Stochastic gradient boosting.
\newblock {\em Computational Statistics \& Data Analysis}, 38(4):367--378,
  2002.

\bibitem{duchi2009boosting}
John Duchi and Yoram Singer.
\newblock Boosting with structural sparsity.
\newblock In {\em Proceedings of the 26th Annual International Conference on
  Machine Learning}, pages 297--304. ACM, 2009.

\bibitem{lugosi2004bayes}
G{\'a}bor Lugosi and Nicolas Vayatis.
\newblock On the bayes-risk consistency of regularized boosting methods.
\newblock {\em Annals of Statistics}, pages 30--55, 2004.

\bibitem{xi2009speed}
Yongxin~T Xi, Zhen~J Xiang, Peter~J Ramadge, and Robert~E Schapire.
\newblock Speed and sparsity of regularized boosting.
\newblock In {\em International Conference on Artificial Intelligence and
  Statistics}, pages 615--622, 2009.

\bibitem{shen2010totally}
Chunhua Shen, Hanxi Li, and Nick Barnes.
\newblock Totally corrective boosting for regularized risk minimization.
\newblock {\em arXiv preprint arXiv:1008.5188}, 2010.

\bibitem{hastie2009elements}
Trevor Hastie, Robert Tibshirani, Jerome Friedman, T~Hastie, J~Friedman, and
  R~Tibshirani.
\newblock {\em The elements of statistical learning}, volume~2.
\newblock Springer, 2009.

\bibitem{tibshirani1996regression}
Robert Tibshirani.
\newblock Regression shrinkage and selection via the lasso.
\newblock {\em Journal of the Royal Statistical Society. Series B
  (Methodological)}, pages 267--288, 1996.

\bibitem{osborne2000lasso}
Michael~R Osborne, Brett Presnell, and Berwin~A Turlach.
\newblock On the lasso and its dual.
\newblock {\em Journal of Computational and Graphical statistics},
  9(2):319--337, 2000.

\bibitem{shalev2010trading}
Shai Shalev-Shwartz, Nathan Srebro, and Tong Zhang.
\newblock Trading accuracy for sparsity in optimization problems with sparsity
  constraints.
\newblock {\em SIAM Journal on Optimization}, 20(6):2807--2832, 2010.

\bibitem{zhao2004boosted}
Peng Zhao and Bin Yu.
\newblock Boosted lasso.
\newblock Technical report, DTIC Document, 2004.

\bibitem{frank1956algorithm}
Marguerite Frank and Philip Wolfe.
\newblock An algorithm for quadratic programming.
\newblock {\em Naval research logistics quarterly}, 3(1-2):95--110, 1956.

\bibitem{jaggi2013revisiting}
Martin Jaggi.
\newblock Revisiting frank-wolfe: Projection-free sparse convex optimization.
\newblock In {\em Proceedings of the 30th International Conference on Machine
  Learning (ICML-13)}, pages 427--435, 2013.

\bibitem{guelat1986some}
Jacques Gu{\'e}lat and Patrice Marcotte.
\newblock Some comments on wolfe's ‘away step’.
\newblock {\em Mathematical Programming}, 35(1):110--119, 1986.

\bibitem{clarkson2010coresets}
Kenneth~L Clarkson.
\newblock Coresets, sparse greedy approximation, and the frank-wolfe algorithm.
\newblock {\em ACM Transactions on Algorithms (TALG)}, 6(4):63, 2010.

\bibitem{koltchinskii2002empirical}
Vladimir Koltchinskii and Dmitry Panchenko.
\newblock Empirical margin distributions and bounding the generalization error
  of combined classifiers.
\newblock {\em Annals of Statistics}, pages 1--50, 2002.

\bibitem{mohri2012foundations}
Mehryar Mohri, Afshin Rostamizadeh, and Ameet Talwalkar.
\newblock {\em Foundations of machine learning}.
\newblock MIT press, 2012.

\bibitem{schapire2012boosting}
Robert~E Schapire and Yoav Freund.
\newblock {\em Boosting: Foundations and algorithms}.
\newblock MIT press, 2012.

\bibitem{slides}
Paul Grigas, Robert Freund, and Rahul Mazumder.
\newblock The frank-wolfe algorithm: New results, and connections to
  statistical boosting.
\newblock Workshop on Optimization and Big Data, University of Edinburgh, 2013.

\bibitem{Lichman:2013}
M.~Lichman.
\newblock {UCI} machine learning repository, 2013.

\end{thebibliography}


\begin{thebibliography}{1}

\bibitem{jaggi2013revisiting}
Martin Jaggi.
\newblock Revisiting frank-wolfe: Projection-free sparse convex optimization.
\newblock In {\em Proceedings of the 30th International Conference on Machine
  Learning (ICML-13)}, pages 427--435, 2013.

\bibitem{guelat1986some}
Jacques Gu{\'e}lat and Patrice Marcotte.
\newblock Some comments on wolfe's ‘away step’.
\newblock {\em Mathematical Programming}, 35(1):110--119, 1986.

\end{thebibliography}
\bibliographystyle{unsrt}
\end{document}